\documentclass[10pt,twocolumn,letterpaper]{article}

\usepackage[pagenumbers]{wacv} 
\definecolor{wacvblue}{rgb}{0.21,0.49,0.74}
\usepackage[pagebackref,breaklinks,colorlinks,allcolors=wacvblue]{hyperref}

\title{
Object-Centric Data Synthesis for Category-level Object Detection
}
\author{
    Vikhyat Agarwal$^{1}$\thanks{Equal contribution.} \quad
    Jiayi Cora Guo$^{2}$\footnotemark[1] \quad
    Declan Hoban$^{3}$\footnotemark[1] \quad
    Sissi Zhang$^{4}$\footnotemark[1] \quad \\[5pt]
    Nicholas Moran$^{5}$ \quad
    Peter Cho$^{5}$ \quad
    Srilakshmi Pattabiraman$^{5}$ \quad
    Shantanu Joshi$^{2}$
    \\
    \\
    $^{1}$University of Richmond \quad
    $^{2}$University of California, Los Angeles \\[3pt]
    $^{3}$University of California, Berkeley \quad
    $^{4}$University of Texas at Austin \quad
    $^{5}$Analog Devices, Inc.
}

\begin{document}
\maketitle
\begin{abstract}

Deep learning approaches to object detection have achieved reliable detection of specific object classes in images. However, extending a model’s detection capability to new object classes requires large amounts of annotated training data, which is costly and time-consuming to acquire, especially for long-tailed classes with insufficient representation in existing datasets. Here, we introduce the object-centric data setting, when limited data is available in the form of object-centric data (multi-view images or 3D models), and systematically evaluate the performance of four different data synthesis methods to finetune object detection models on novel object categories in this setting. The approaches are based on simple image processing techniques, 3D rendering, and image diffusion models, and use object-centric data to synthesize realistic, cluttered images with varying contextual coherence and complexity. We assess how these methods enable models to achieve category-level generalization in real-world data, and demonstrate significant performance boosts within this data-constrained experimental setting. Code for this work is available at \url{https://github.com/RIPS25-Analog/OC-Synthesis}.
\end{abstract}
    
\section{Introduction}
\label{sec:intro}

Object detection is a computer vision task with widespread applications in diverse areas like healthcare, agriculture, and autonomous vehicles \cite{he2025odmed, badgujar2024agri, balasubramaniam2022autov}. Deep learning models offer fast and robust detection of different object classes in real-world images, but a significant obstacle to their practical adoption is the lack of high-quality labeled training data,  especially when dealing with rare object classes that are not well-represented in existing image datasets.

Humans can very quickly learn to identify a new type of object even after only seeing a few examples of that object in the same background setting. In contrast, modern object detection models require substantial amounts of annotated data to achieve robust performance in a range of visual conditions. Models trained on too little data tend to overfit and struggle to generalize to novel complex scenes \cite{antonelli2022few}. For example, the creation of the COCO data set required tens of thousands of hours of manual human effort to annotate object positions and segmentation masks \cite{COCO,remez2018learning}. This presents a significant time and cost barrier when scaling object detection systems to new object categories. Thus, one approach to managing data scarcity is to augment the existing real dataset with synthetic data, i.e. images that can be generated algorithmically. These images can be generated in large quantities at relatively low cost and include precise annotations by design. Moreover, synthetic generation processes can allow direct control over parameters such as object pose, lighting, occlusion, and background composition. This flexibility enables the construction of more balanced and diverse datasets, which are especially valuable when labeled real-world data is limited or costly to collect. When thoughtfully integrated, synthetic data has been shown to improve both the robustness and generalization of object detection models \cite{ozeren2025evaluating}. 

A common setting for object detection under data scarcity is few-shot object detection (FSOD), where one must learn to detect new object classes given only a few example images of each of these classes \cite{huang2022survey-ssod-fsod}. In this paper, we identify another setting of data scarcity where we must learn to detect novel object classes given object-centric data (\eg multi-view images or 3D models of object instances). This setting is similar to practical applications where one has physical access to some instances of the novel classes (\eg \cite{shamsafar2023leveraging}), rather than having a small number of images of the objects placed in different scenes. In light of automatic segmentation tools (e.g. Meta's SAM-2 with video tracking \cite{ravi2024sam2}) or modern photogrammetry software (e.g. Epic Games' RealityScan \cite{RealityScan_2025}), we note that obtaining masked images and 3D models of isolated objects is feasible and relatively automatable.

We also emphasize that existing works often fail to address key considerations of practical applications, due to (\textbf{1}) insufficient modeling of occlusion or cluttered environments and (\textbf{2}) dependence on generative models, which cannot reliably synthesize long-tail object categories that are unrepresented in image datasets (e.g. specialized industrial or medical equipment), as noted by \cite{zhu2024odgen}. Moreover, previous works have reported mixed findings regarding the effectiveness of incorporating extra visual context in synthetic images \cite{dvornik2018modeling, Ghiasi2020SimpleCI}. In this work, we develop data synthesis methods of varying levels of contextual coherence that address limitations \textbf{1} and \textbf{2}, and we compare these methods in the particular data-limited setting where we just have access to object-centric data (complete, isolated views of objects, e.g. 3D models and multi-view images) instead of expensively annotated data of the objects placed in various environments. We propose the following contributions in this paper:

\begin{itemize}[leftmargin=8mm] 
    \item We perform a direct comparative assessment of different data synthesis methods, each incorporating different levels of information about the target objects and their visual context, in a specific limited-data setting that is relevant to practical applications but under-addressed.
    \item We choose (or develop) methods which model occlusions and cluttered environments, and which work on long-tailed object classes. We propose and implement two new data synthesis methods, Diffusion Copy-Paste and 3D Random Placement, both operating on object-centric data.
\end{itemize}
\section{Related Work}
\label{sec:background}

\subsection{Synthetic Data Generation}\label{subsec:synthdata}
Synthetic data generation is a popular approach for expanding machine learning training datasets in limited-data scenarios, and much prior work has demonstrated its efficacy in different domains \cite{figueira2022-survey-synthetic-data-GANs}. Popular methods for synthetic image generation include image composition using classical image processing, 3D rendering engines, and generative deep-learning approaches \cite{man2022-synthetic-image-data}. 

Traditional synthetic data methods for object detection relied on techniques such as geometric image deformations like translation and shears (\eg \cite{lecun2002gradient}) or photometric transformations like color jittering and lighting perturbation \cite{mumuni2024-survey-synthetic-data-augmentation}. These augmentations create visually diverse variants of the existing data, but cannot alter the underlying image scene, giving limited control over diversity of scene composition.

To synthesize novel training instances from limited data while controlling scene-level parameters such as object positioning, \cite{CutNPaste-Dwibedi2017} presented the Cut-Paste pipeline. Cut-Paste relies on image composition: starting with a set of masked foreground objects, pasting them onto various background images in random orientations and applying {\lq blending\rq} at the object boundaries. The blending ensures local or {\lq patch-level\rq} realism in the composite images, so an object detection model trained on this data does not fixate on boundary-level discrepancies that only appear in the synthetic data. Additionally, the authors recommend using {\lq distractors\rq}: objects pasted onto images that do not belong to any of the target classes. These make the basic object recognition task tougher, since models can no longer identify target class objects in the image by flagging any objects that look {\lq out of place\rq}. This simple approach is found to be effective in training object detection models when used alongside real annotated data.

Approaches inspired by Cut-Paste use background context to place objects in sensible positions, hence imposing global realism in the synthetic samples. Concurrent to Cut-Paste, \cite{georgakis2017synthesizingtrainingdataobject} performs plane detection in indoor RGBD backgrounds to find physically plausible positions for pasting objects, and noted performance improvements over random Cut-Paste in certain settings. \cite{dvornik2018modeling} presented a CNN {\lq context model\rq} trained in self-supervised manner to suggest contextually sensible object placements. They found that random Cut-Paste \cite{CutNPaste-Dwibedi2017} performs worse than using traditional data augmentation techniques, but their contextual augmentation yields significant improvement. Their experimental setting of category-level object detection differs from the instance-level setting of \cite{CutNPaste-Dwibedi2017}, where different objects of the same category are marked as unique labels (\eg \lq Cup 1\rq, \lq Cup 2\rq) rather than being treated as the same category-level label (\lq Cup\rq). The explanation offered by \cite{dvornik2018modeling} is that instance detection requires a model to learn fine-grain appearance of an object instance, while general object detection requires generalizing to unseen instances of the same category,  which may benefit from more context modeling to provide category-relevant cues. \cite{fang2019instaboost} decides object placement based on local contour similarity in the background image, and they report improvements over random Cut-Paste \cite{CutNPaste-Dwibedi2017} and contextual Cut-Paste \cite{dvornik2018modeling} in the task of instance segmentation.

In the interest of reducing the synthetic-to-real domain gap, previous works have explored the use of 3D computer graphics for image generation. Rendering a complete physical scene in $3$ dimensions is critical when a complete understanding of the object's position, orientation, and scale is needed, such as for 3D object detection tasks \cite{3D-copy-paste}. However, even for the simpler problem of detecting objects in an image, 3D rendering can supply annotated training data that is more visually coherent, especially when using photorealistic rendering \cite{Hodan2019PhotorealisticIS, useful-photorealistic-render}. For example, \cite{synthetica} used NVIDIA's Omniverse Isaac Sim \cite{isaacsim} to create a pipeline for large-scale synthetic image generation using photorealistic ray-tracing, rendering randomization, and physics simulation, to achieve robust real-time instance detection.

Finally, the use of generative AI modeling for data synthesis has gained popularity in recent years. Earlier techniques using Generative Adversarial Networks (GANs) and conditional GANs enabled the creation of completely new images of target objects from a text or image prompt \cite{bowles2018gan, cycleGAN}. Recently many methods have made use of diffusion models with different approaches, including generative dataset expansion guided by semantic similarity \cite{zhang2023expanding}, or augmenting existing object detection datasets with labeled generative images \cite{zhang2025diffusionengine}. Some more targeted methods include InstaGen's generation of diffusion images with annotations \cite{feng2024instagen}, ODGEN's image generation conditioned on provided bounding boxes annotations \cite{zhu2024odgen}, or X-Paste's generation of foreground target objects that are then pasted onto various backgrounds \cite{zhao2023x}. The relevance of these different data synthesis techniques in our experimental setting will be elaborated in the next section.

\subsection{Data-Scarce Scenarios}\label{subsec:FSOD}

There exist other approaches to data-limited object detection besides synthetic data generation, such as model-oriented approaches that adapt existing object detection model architectures to the few-shot context by efficiently separating different target classes in feature space (see \cite{liu2023recent}). Though we make use of transfer learning --- by finetuning a pretrained object detection model on data with novel object classes ---- we are primarily concerned with synthetic data-oriented approaches to alleviate data scarcity. These offer model-agnostic improvements which can be carried over as new and improved object detection architectures develop rapidly in the computer vision literature. These data-oriented methods also avoid inference-time overhead and require only increased training compute. Crucially, our setting differs from few-shot object detection by assuming access to isolated captures or 3D models of the targets. This models the practical scenario where a user has specific physical examples of the target categories they want to detect and distinguish between, rather than annotated data of the objects placed in diverse environments.

Furthermore, we are concerned with the case of long-tail target classes which may not be represented in existing image datasets (such as those on which an image generation model would be trained). To elaborate on this distinction and justify our choice of data synthesis methods, we may contrast our selected methods with those found in the existing literature. For example, \cite{zhao2023x} introduce X-Paste, a modification of the Cut-Paste pipeline which uses Stable Diffusion \cite{rombach2022high} and CLIP \cite{radford2021learning} to generate and filter images of target foreground objects to be pasted onto background images. \cite{lin2023explore} applies a similar approach to the FSOD task, but still relies on pretrained diffusion models for target object generation. While these approaches improve the scalability and performance of traditional Cut-Paste by synthesizing new object examples, they are limited in their application to target classes which can reliably be generated by a diffusion model. Diffusion models struggle to reliably generate diverse images of rare objects that lack significant representation in their training set \cite{samuel2024generatingRare}, and this poses a significant issue if our object class of interest is, for \eg, specialized industrial or medical equipment, where errors in minor details can lead to catastrophic misclassifications. Other methods using feature extractors \cite{liu2025control} or controllable diffusion \cite{fang2024data} are able to retain certain visual characteristics of generated foreground objects without training on large-scale object detection datasets, but still rely on pre-trained diffusion models to fill in target object details or fail to model occlusions between objects, as noted in \cite{zhu2024odgen}. In contrast to the previous approaches, our proposed data augmentation technique based on conditional diffusion allows for precise control over object positions and occlusion levels; does not substantially alter the appearances of the target classes or require them to be represented in existing datasets; and can still condition background generation on an image prompt (see also \cite{li2025domain}).

In summary, the control over target class positions, occlusion levels, and visual appearances of target objects, as well as the lack of reliance on existing large-scale datasets, uniquely characterize our chosen data augmentation approaches described in \cref{sec:syntheticdata}: Cut-Paste \cite{CutNPaste-Dwibedi2017}, Diffusion Copy-Paste, 3D Random Placement, and 3D Copy-Paste \cite{3D-copy-paste}.
\section{Data Synthesis Methods}\label{sec:syntheticdata}

We select data synthesis methods which require either a set of masked images of target objects (for 2D Cut-Paste or Diffusion Copy-Paste) which can be generated using automatic segmentation tools like SAM2 \cite{ravi2024sam2}, or 3D models of target and distractor objects (for 3D Random Placement and 3D Copy-Paste) which can be generated with automatic photogrammetry tools (\eg \cite{RealityScan_2025}). Moreover, our chosen methods cover a diverse range of image processing and generation techniques and incorporate varying degrees of contextual coherency of target objects in the synthesized images. We are particularly interested in understanding how incorporating synthetic images derived from object-centric data can improve object detection in our experimental settings.

\subsection{Cut-Paste}\label{subsec:2d-cut-paste}

\begin{figure}[htp]
    \centering
    \includegraphics[width=\linewidth]{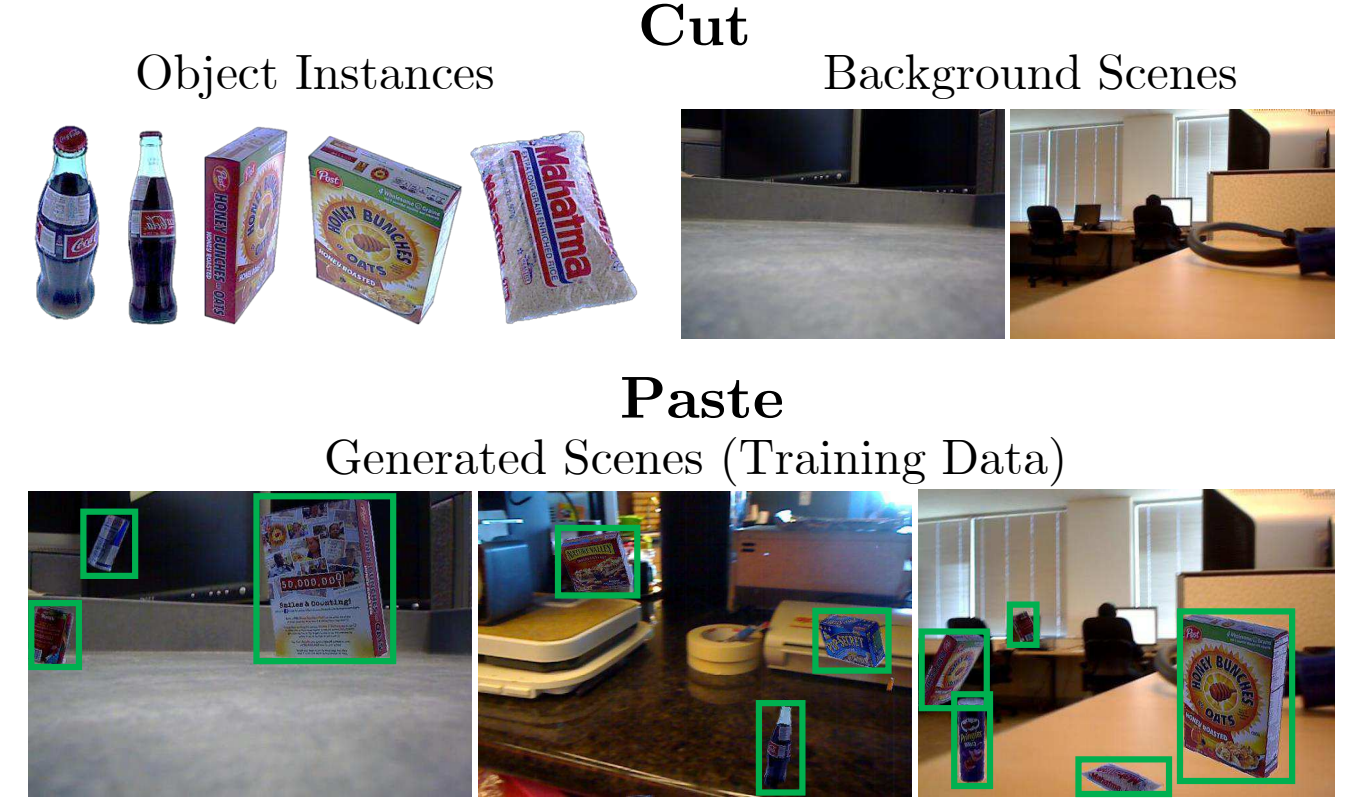}
    \caption{Examples from the Cut-Paste implementation of \cite{CutNPaste-Dwibedi2017}.}
    \label{fig:CutnPaste}
\end{figure}

As described in \cref{subsec:synthdata}, the Cut-Paste method \cite{CutNPaste-Dwibedi2017} is a straightforward yet effective approach to synthetic data generation. The process involves extracting object instances from existing images using segmentation masks and then compositing them onto new background scenes (see \cref{fig:CutnPaste}). To ensure variety, the method includes augmentations such as rotation, scaling, occlusion, truncation, and different blending strategies. These additions help create diverse training examples that are both realistic and challenging for object detectors.

One of the key benefits of this method is its ability to produce high-coverage datasets with minimal human intervention. Since object masks and backgrounds can be reused in multiple combinations, the approach significantly increases visual diversity across backgrounds and scales. The method also bypasses the need for precise scene layout or scene lighting information, which are often difficult to replicate or transfer between domains. As a result, it becomes easier to build datasets tailored to specific target environments or rare object instances. We adapt the original implementation (made available by the authors at \url{https://github.com/debidatta/syndata-generation}), porting from Python 2 to 3, adding proportional object scaling, improving the occlusion checks by using object masks, and switching the Poisson Blur effect to OpenCV's Seamless Cloning \cite{opencv_library} for speed.

\subsection{Diffusion Copy-Paste}\label{subsec:diffusion-copy-paste}
We propose a novel method using generative models for synthetic data generation. This approach, termed as {\em Diffusion Copy-Paste} uses diffusion models conditioned on certain edge and image prompts. In situations where it is important to maintain the visual appearance of target objects found in training data (such as for long-tailed or sensitive target classes), one can control the distribution of generated images using conditional diffusion tools. For example, we use ControlNet \cite{zhang2023adding}, a neural network architecture for adding spatial conditioning controls (in our case, edge map conditioning) to large, pretrained text-to-image diffusion models. Additionally we use an Image Prompt Adapter \cite{ye2023ip-adapter}, which conditions the background of the synthesized image to match a reference background image. Given pictures of the environment one wants to perform object detection in, Diffusion Copy-Paste can use these by passing them in to the Image Prompt Adapter, hence creating images which have similar backgrounds to the target environment.

\begin{figure}[ht]
    \centering
    \includegraphics[width=\linewidth]{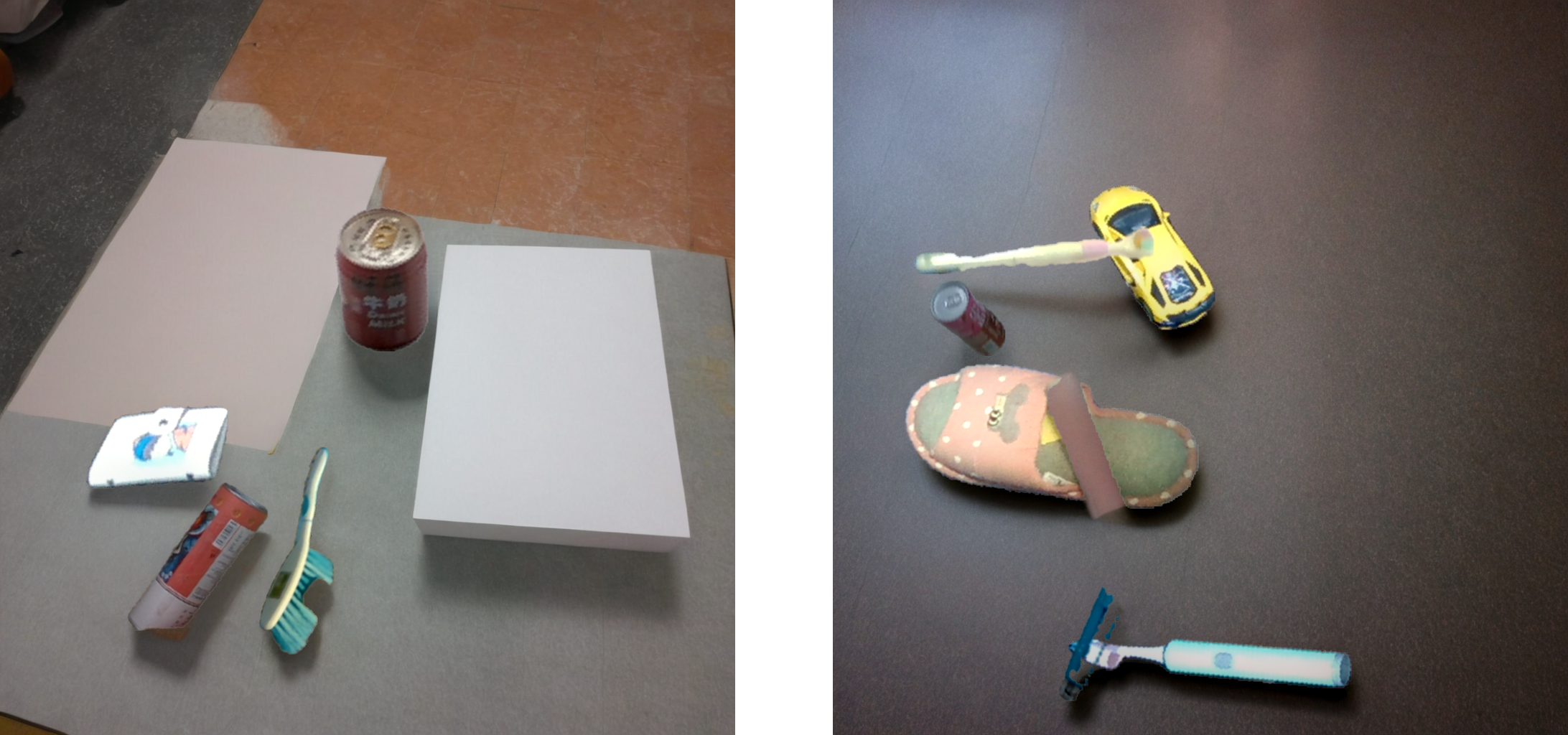}    \caption{Some examples of the Diffusion Copy-Paste image generation technique.}
    \label{fig:examples_diff}
\end{figure}

\begin{figure}[ht]
    \centering
    \includegraphics[width=\linewidth]{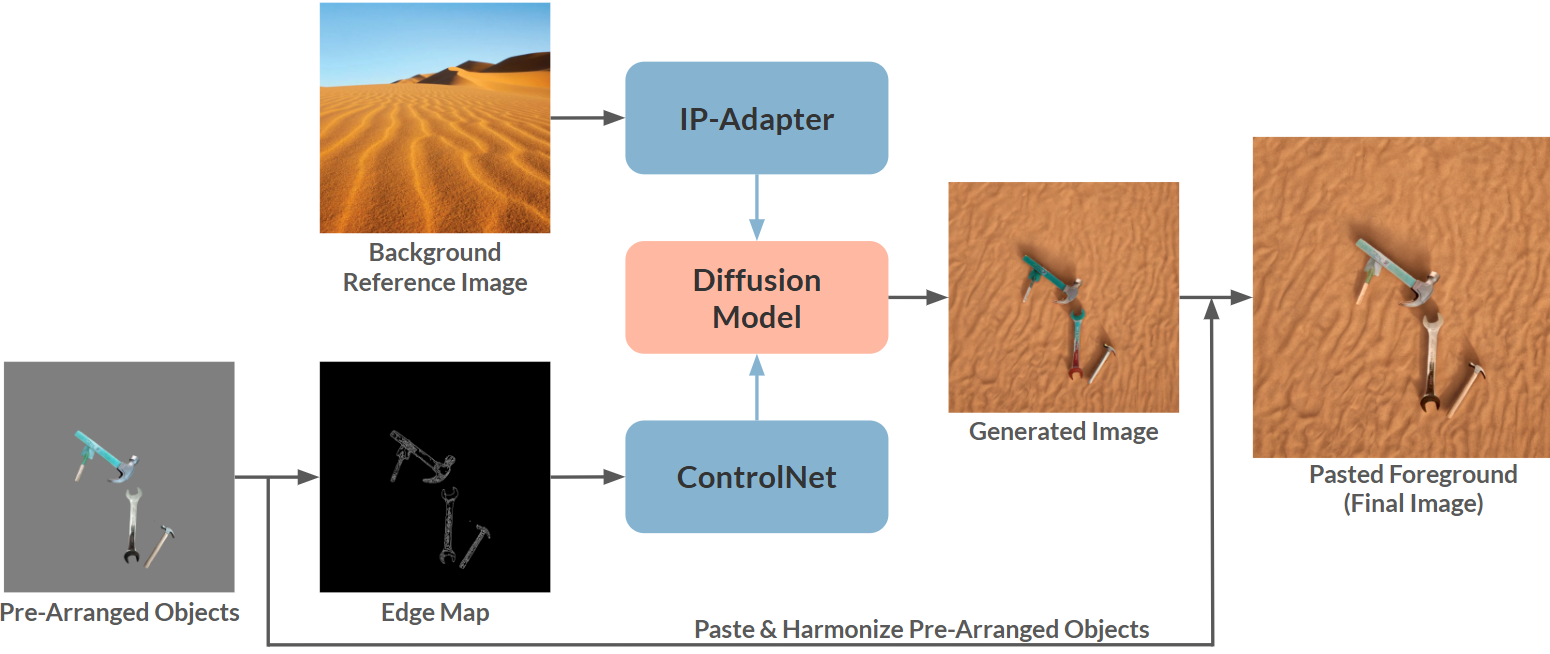}    \caption{Pipeline for synthetic data generation for Diffusion Copy-Paste.}
    \label{fig:pipeline}
\end{figure}

\cref{fig:pipeline} demonstrates the complete Diffusion Copy-Paste pipeline: \textbf{1)} pre-arranged objects (obtained by running Cut-Paste with a blank background) are passed into the Canny edge detector algorithm \cite{canny} to obtain an edge map that encodes their positions and outlines within the picture. \textbf{2)} we generate a synthetic image conditioned to follow a reference background image prompt (through IP-Adapter) and have edges matching the object layout edge map (through ControlNet). Since the foreground objects generated by the diffusion model are often nonsensical, we then paste the original pre-arranged objects and perform basic image harmonization to obtain the final image.

The implementation of the diffusion method expands on the 2D Cut \& Paste method but also accesses certain core and custom nodes from the ComfyUI backbone \cite{ComfyUI_Official_Documentation_2024} for running the image generation model and image post-processing tasks. The specific image generation model used was RealVisXL v5.0 Lightning \cite{realvisxl-huggingface}, a diffusion model optimized for fast photorealistic image generation.

\subsection{3D Random Placement}\label{subsec:3D-RP}
\begin{figure}[ht]
    \centering
    \includegraphics[width=1\linewidth]{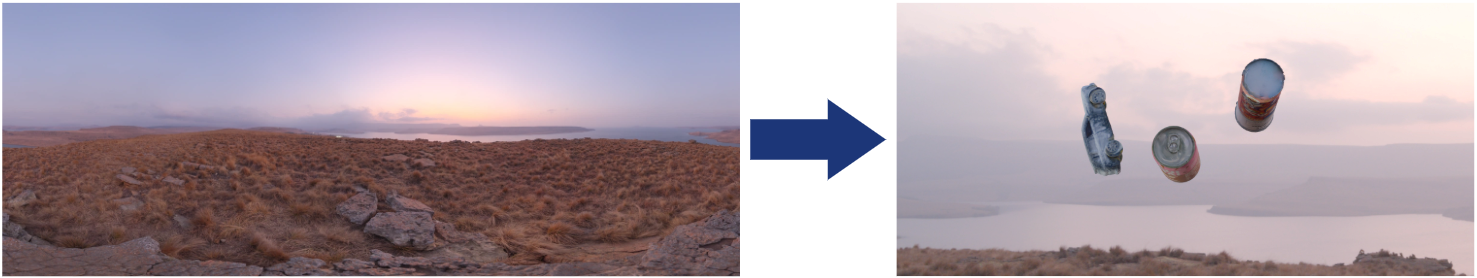}
    \caption{From HDRI background to synthetic data.}
    \label{fig:3d-rp}
\end{figure}

We also propose a method for utilizing the 3D geometry of the objects by rendering 3D scenes with objects randomly placed in them. This method is referred to as {\em 3D Random Placement (3D RP)} and employs high dynamic range image maps (HDRI)  as backgrounds. HDRI scenes capture complete 360° views of real environments while preserving lighting intensity on each image pixel. This property ensures the 3D scene has consistent and realistic lighting without the need for added light sources, thereby allowing foreground objects to blend in naturally with the background (as seen in \cref{fig:3d-rp}). In this approach, objects are still inserted randomly into the background, leading to objects that float in mid-air and have no contextual relation to the background. 

\begin{figure}[ht]
    \centering
    \includegraphics[width=1\linewidth]{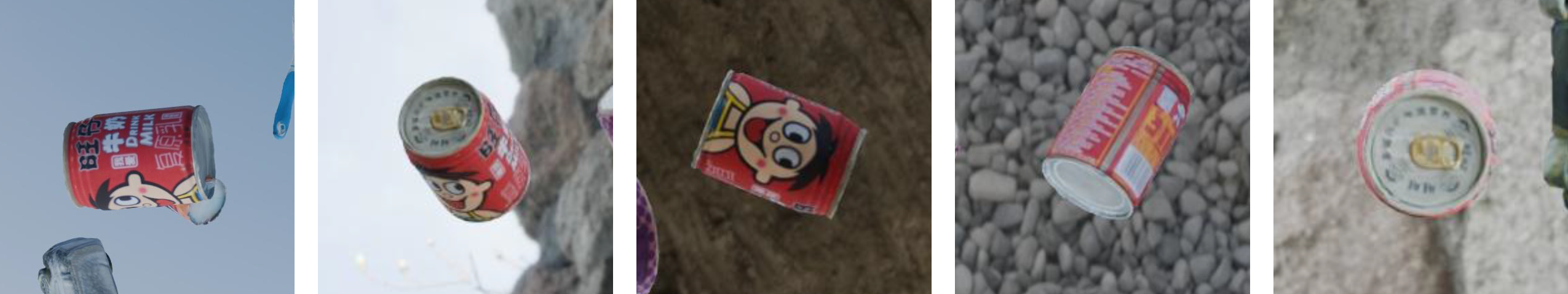}
    \caption{A single can viewed from multiple angles.}
    \label{fig:multiple-views}
\end{figure}

\noindent We arrange multiple 3D scenes in the Blender rendering engine \cite{blender} (using Python bindings), each with multiple target \& distractor objects placed in different positions. 3D RP renders each scene multiple times with varying camera angles, zoom levels, and exposure levels, to capture diverse object viewpoints (as seen from \cref{fig:multiple-views}) and model the objects' appearances under various lighting conditions.

\subsection{3D Copy-Paste}\label{subsec:3D-CP}
Adapted from \cite{3D-copy-paste}, this method adds contextual coherence to the 3D Random Placement method by placing objects in a collision-free manner on feasible horizontal planes, which are detected within the RGBD background images using clustering algorithms. The two primary improvements of 3D Copy-Paste over 3D Random Placement are physically plausible object placements (using plane detection) and lighting estimation (using deep inverse rendering techniques \cite{li2020inverse}).

The original implementation utilizes two primary computational steps: geometric analysis of scenes and spatially-varying illumination estimation. Agglomerative Hierarchical Clustering \cite{feng14_AHC} identifies planar surfaces from RGBD point clouds, these are narrowed to horizontal planes by comparing the plane surface normals to the z-axis, and the {\lq floor\rq} plane is chosen as the lowest horizontal plane. The scene's lighting is estimated using deep inverse rendering through LightNet \cite{li2020inverse}, which predicts environment maps of omnidirectional light distribution over discrete $4 \times 4$ pixel areas. 

\begin{figure}[ht!]
    \centering
    \includegraphics[width=1\linewidth]{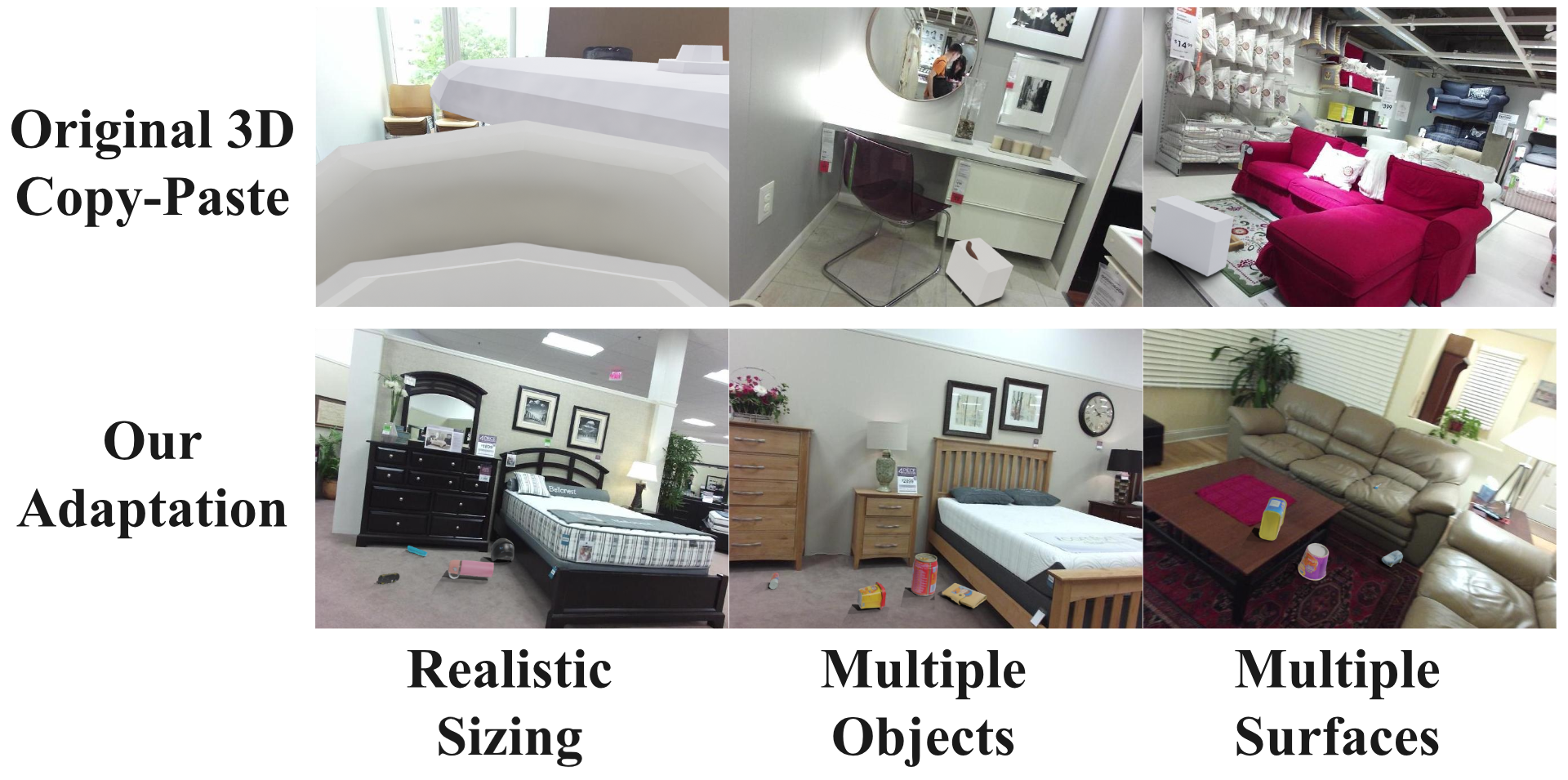}
    \caption{Our improvements on the original 3D Copy-Paste method \cite{3D-copy-paste}.}
    \label{fig:3d_copy_paste_improvements}
\end{figure}

Our adaptation of 3D Copy-Paste involved five new features: (1) allowing objects to be inserted on multiple horizontal surfaces instead of just the floor plane. (2) inserting multiple objects in a scene (we sequentially search for free space and insert objects one-by-one). (3) realistic object sizing (we detect and prevent scenarios where objects take up too much space in the camera's view). (4) rotating objects along the $x,y,z$ axes, rather than $z$-axis only (we remove the assumption that objects have a {\lq base\rq} which always rests on the floor). (5) 3D collision detection (this replaces the original 2D approximations with precise 3D bounding box collision checks). \Cref{fig:3d_copy_paste_improvements} summarizes our improvements.

\subsection{Comparison}\label{subsec:comparison}
\begin{figure}[ht]
    \centering
    \includegraphics[width=\linewidth]{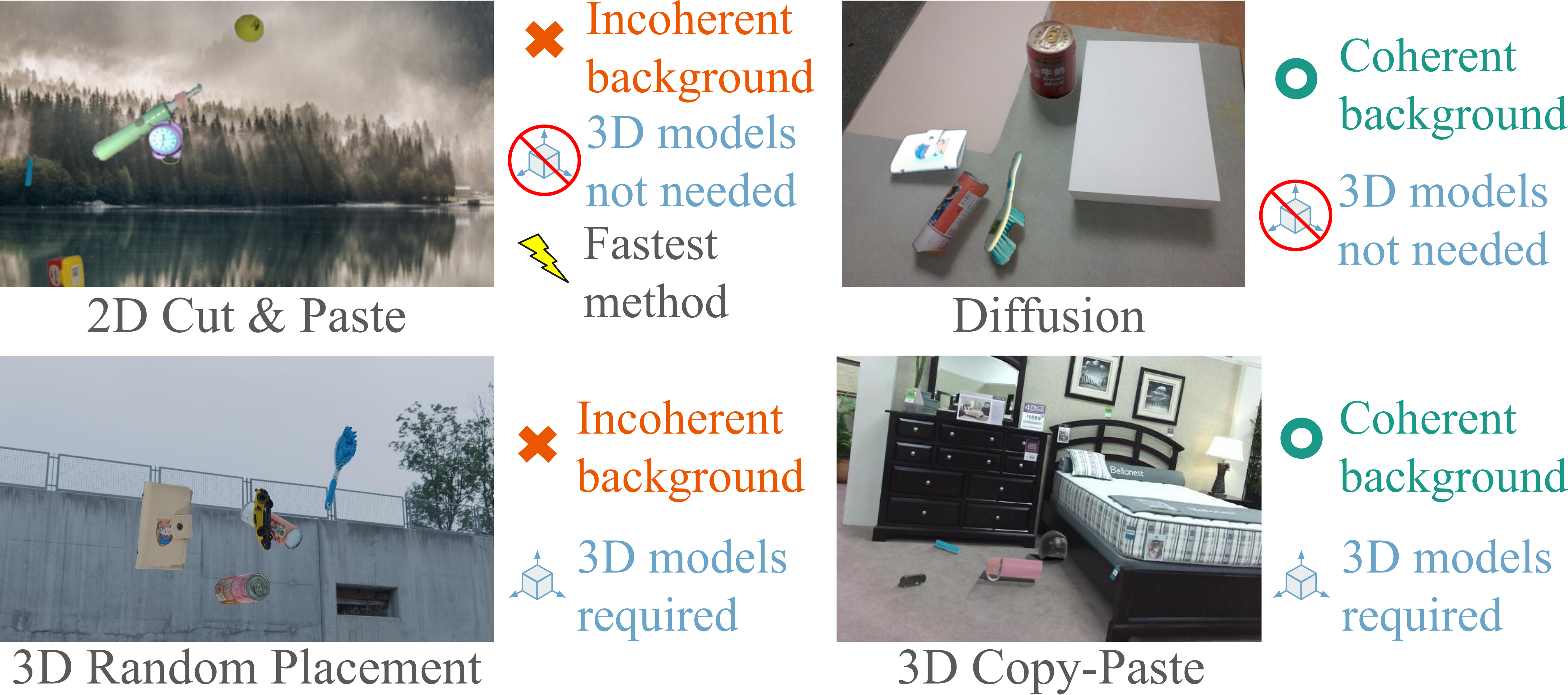}    \caption{Qualitative comparison of our proposed data generation methods.}
    \label{fig:comparison}
\end{figure}

The four data synthesis methods vary in the input data types they utilize and they incorporate different amounts of visual context into the images they  synthesize. While the objects in 2D Cut-Paste and 3D Random Placement have almost no relation to the backgrounds, the Diffusion Copy-Paste and 3D Copy-Paste methods use contextual information about the foreground objects and backgrounds to produce a more coherent and realistic object placement. 3D rendering of objects and incorporation of background context comes at a computational cost, however, with 2D Cut-Paste producing synthetic images at about 10 times the rate (25 imgs/sec) as the other three methods, though all of the synthetic data methods are still far cheaper to execute than real annotated data collection. \Cref{fig:comparison} contains example images generated by these methods and an overview of their benefits and drawbacks.

\section{Evaluation Dataset}\label{sec:datasets}

To evaluate our different methods of generating synthetic data, we would like to train object detection models on these datasets and evaluate these models on real-world data. As described in the previous sections, we are particularly interested in situations where labeled training data containing target objects may not be available in large amounts, but object-centric data is available to us. In light of this, we select the PACE dataset \cite{you2024pace} as our source of real-world evaluation data. PACE consists of $300$ videos with $55,000$ frames and $258,000$ object annotations, covering $238$ object instances from $43$ categories in cluttered scenes. The authors also provide 3D textured models for each object instance. The dataset is accompanied by PACE-Sim, which uses BlenderProc's physically-based rendering \cite{Denninger2023-blenderproc2} to create $100,000$ synthetic images with $2.4$ million annotations across $931$ object instances, but we discard this since it was generated using similar methods to 3D RP and 3D Copy-Paste (which use the Blender engine), making it unsuitable for the purpose of evaluation.

Out of the $43$ object categories represented in PACE, $13$ are either represented in COCO \cite{COCO} or closely related to classes in COCO (\eg PACE has mugs while COCO has cups). We discard these classes since our experiments use the YOLO11 \cite{yolo11_ultralytics} and RT-DETR \cite{zhao2024_rtdetr} object detection models, which were pretrained on the COCO dataset, hence giving them out-of-the-box detection capabilities on these. Out of the remaining $30$ classes, we discard objects with fewer than $10$ unique instances appearing in the real videos, and choose our final object classes to be ``toy car", ``can", ``snack box", and ``ramen box".

To set up an experimental setting which tests generalization across different instances of the same category, we split the instances for each class into training, validation, and test splits in a $2:1:2$ ratio, while ensuring that instances that co-occur in any video are allocated to the same split (this is done by creating an instance co-occurrence graph and using a greedy algorithm to split the components in the desired ratio). This allows us to then assign each video to the same split as all the object instances that appear in the video, resulting in the split described in \Cref{tab:dataset-stats}.

\begin{table}[h]
  \centering
  \begin{tabular}{@{}lccc@{}}
    \toprule
    & Train & Val & Test \\
    \midrule
    Toy Car Instances   & 5 & 2 & 4 \\
    Can Instances       & 4 & 2 & 4 \\
    Snack Box Instances & 7 & 4 & 7 \\
    Ramen Box Instances & 6 & 3 & 5 \\
    \cmidrule(l){2-4}
    \textbf{Total Instances}     & 22 & 11 & 20 \\
    \cmidrule(l){2-4}
    Videos              & 75  & 33  & 75  \\
    Frames              & 11,498 & 5,547 & 10,825 \\
    \bottomrule
  \end{tabular}
  \caption{Statistics of our dataset across train, validation, and test splits, including per-class object instances.}
  \label{tab:dataset-stats}
\end{table}

\section{Experiments and Results}\label{sec:experiments_results}

\subsection{Design Choices}

We generate 20k images for each synthetic data method, using the object-centric data from the PACE dataset. Cut-Paste and Diffusion Copy-Paste use foreground cutouts for the target and distractor instances found in video frames of the training data. As backgrounds, they use PACE  video frames that lie completely outside the train-val-test above, to avoid test leakage or target objects appearing in background without being assigned labels. 3D RP and 3D Copy-Paste use 3D models for the same train instances of target and distractors, and they use external datasets for the backgrounds, namely HDRI images from PolyHaven \cite{PolyHaven} and RGBD images from SUN-RGBD \cite{song2015sun-rgbd}.

We choose the YOLO11 object detection model \cite{yolo11_ultralytics} for our experiments due to its real-time performance \cite{yolo_survey} and the actively-maintained Ultralytics library. YOLO11 was trained on the COCO dataset \cite{COCO} with 80 object classes, but we apply transfer learning by replacing the last layer to classify objects between the 4 categories chosen in our PACE dataset.

\subsection{Sequential versus Mixed schemes}

We first test two different schemes for training the model. In the \textit{sequential} scheme, we first finetune the model on a synthetic dataset only, and then finetune on a fraction of the available PACE training data. In the \textit{mixed} scheme, we combine a synthetic dataset with a fraction of the PACE train data. The sequential and mixed data schemes model the practical scenario where some real annotated data is available (in limited amounts), and synthetic data is used to augment the real data. In the mixed scheme real \& synthetic data are naively combined, while the sequential scheme first finetunes on the abundant synthetic data (to approximately learn the target data distribution) and then finetunes on the limited real data which matches the target data distribution. Previous work \cite{vanherle2022analysis} found the finetuning scheme outperforms the mixed scheme regardless of real data quantity, and we find similar results in our experimental setup, as seen in \Cref{tab:synthetic_v_mixed_results}.


\begin{table}[ht]
    \centering
    \begin{tabular}{l|c|c|c}
         & 2.5\% & 5\% & 10\% \\
        \hline
        Cut-Paste   & 55.6 / 34.8 & 56.1 / 46.2 & 56.3 / 51.1 \\
        Diffusion CP  & 60.5 / 52.1 & 60.2 / 55.3 & 58.4 / 55.0 \\
        3D RP  & \textbf{62.8} / 51.7 & \textbf{64.0} / 52.9 & \textbf{63.2} / 56.9 \\
        3D CP  & 58.2 / 59.8 & 59.6 / 53.5 & 54.7 / 59.7 \\
    \end{tabular}
    \caption{YOLO11 finetuning results (test set mAP@50) with different methods and varying proportions of real data. Inner entries are in the form \textit{Sequential} / \textit{Mixed} performance.}
    \label{tab:synthetic_v_mixed_results}
\end{table}

\subsection{Head-to-Head comparison}
\setlength{\tabcolsep}{16pt}
\begin{table*}[ht!]
    \centering
    \begin{tabular}{l|ccccccc}
         & 200 & 500 & 1,000 & 2,000 & 5,000 & 10,000 & 20,000 \\
         \hline
        Real only & 0.037 & 0.15 & 0.17 & 0.35 & 0.41 &\textbf{ 0.47} & \\
        \hline
        Cut-Paste & 0.35 & 0.34 & 0.34 & 0.32 & 0.38 & 0.39 & 0.33\\
        Diffusion CP & \textbf{0.41} & \textbf{0.38} & \textbf{0.42} & 0.39 & \textbf{0.44} & 0.41 & 0.37\\
        3D RP & 0.38 & 0.38 & 0.40 & \textbf{0.45} & 0.42 & 0.43 & \textbf{0.43}\\
        3D CP & 0.30 & 0.37 & 0.40 & 0.27 & 0.33 & 0.37 & 0.30\\
        \hline
    \end{tabular}
    \caption{Head-to-head comparison of YOLO11 performance (test set mAP@50) with the four different data synthesis methods. The off-the-shelf model is finetuned on the number of images specified by the column, of type specified by the row.}
    \label{tab:head2head}
\end{table*}
\setlength{\tabcolsep}{6pt}
As described in Section \ref{subsec:comparison}, the different data synthesis methods differ in their incorporation of contextual information and in data generation speed. Thus, to directly compare the performance scaling across methods as we increase the number of training images, we performed a head-to-head comparison in which an off-the-shelf YOLO11 model was finetuned on the same number of images of each data type. The results are presented in \Cref{tab:head2head}, and indicate the strong performance of our Diffusion Copy-Paste and 3D Random Placement methods as standalone sources of data, only being surpassed in the scenario where one has 10,000 real images to train on (which exists outside the limited-data regime).

\subsection{Performance across models}
\begin{figure}[ht!]
    \centering
    \includegraphics[width=1\linewidth]{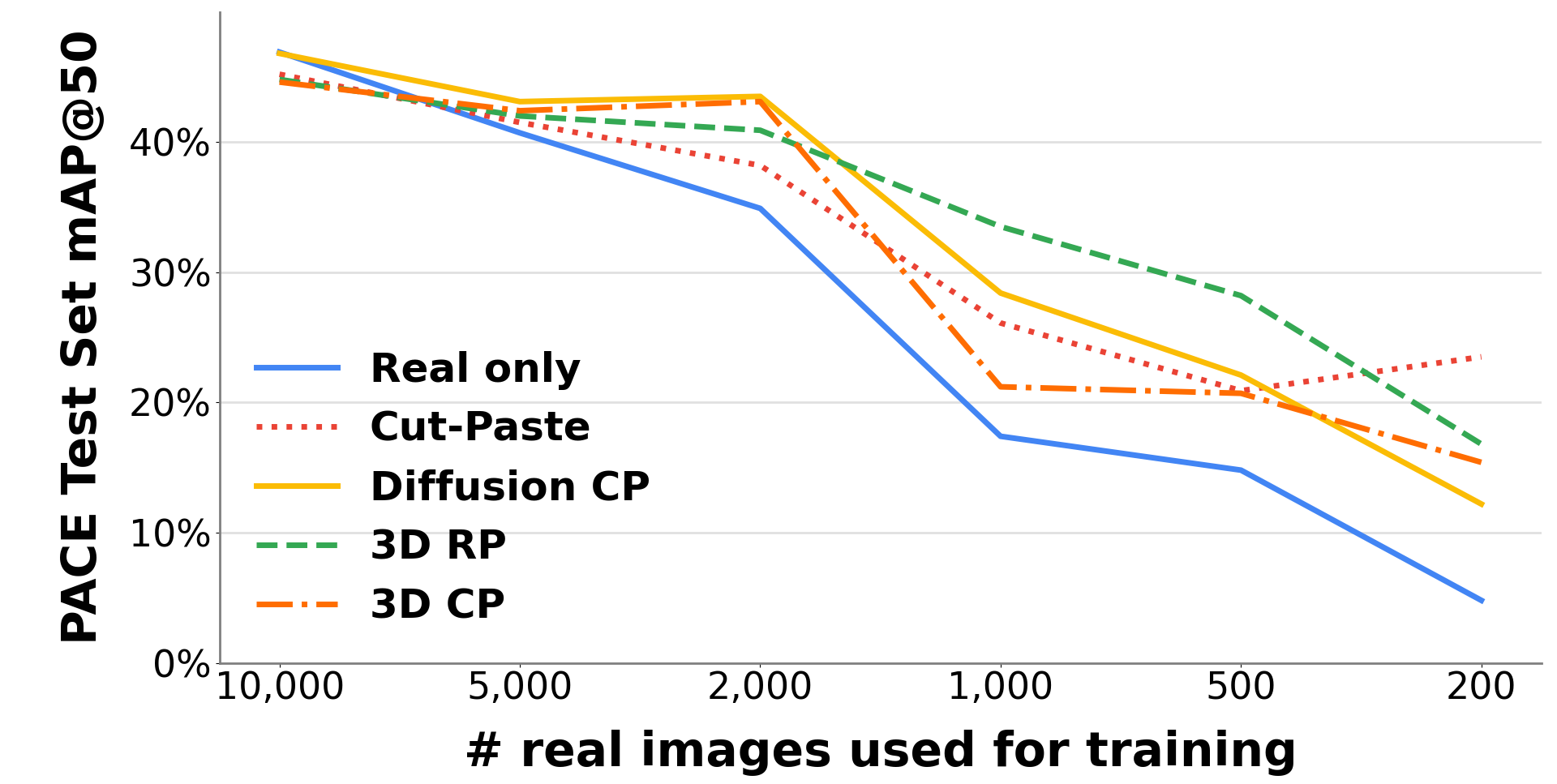}
    \caption{Sequential training of YOLO11 on different amounts of real data, compared to only training on real data.}
    \label{fig:YOLO_seq}
\end{figure}

\begin{figure}[ht!]
    \centering
    \includegraphics[width=1\linewidth]{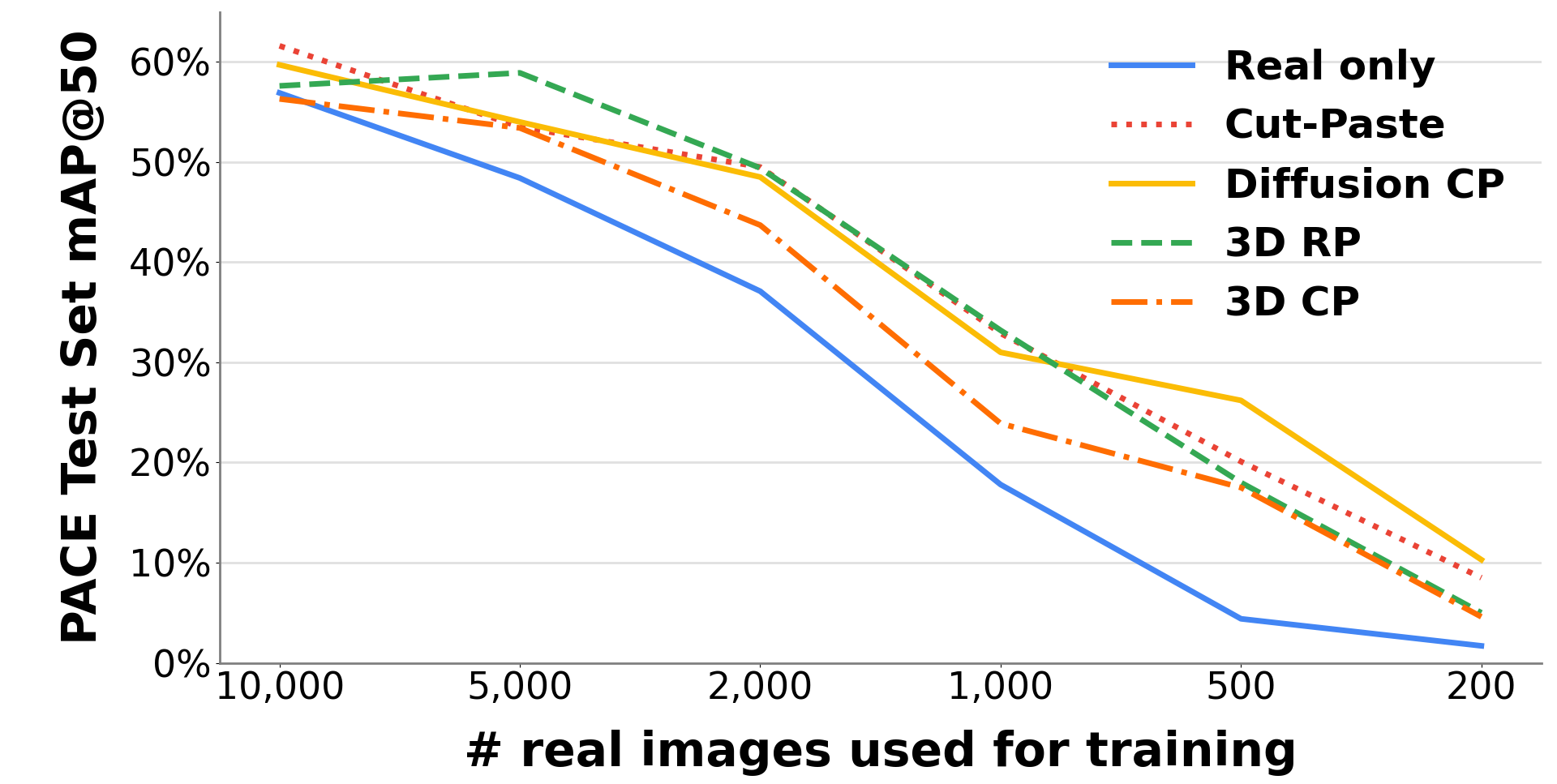}
    \caption{Sequential training of RT-DETR on different amounts of real data, compared to only training on real data. }
    \label{fig:RTDETR_seq}
\end{figure}

We compare the performance of YOLO11 using the sequential training paradigm (\cref{fig:YOLO_seq}) against that of another object detection model, RT-DETR \cite{zhao2024_rtdetr} (\cref{fig:RTDETR_seq}), to investigate if our performance gains hold across different model architectures\footnote{We do not perform hyperparameter optimization in this experiment due to the lack of a consistent validation set when varying the availability of real images.}. We observe the same significant mAP boost from our data augmentation methods, again with strong performances from both 3D Random Placement and Diffusion Copy-Paste. For either model we observe that performance from sequential and  real-only training schemes eventually converge (when 10,000 real images are available for training) but the performance gap between them is substantial in the low-data regime. This shows that a model finetuned on synthetic data offers significant improvement over a pretrained model, when followed by finetuning on a small amount of real data.


\subsection{Performance across occlusion levels}
\begin{figure}[ht!]
    \centering
    \includegraphics[width=1\linewidth]{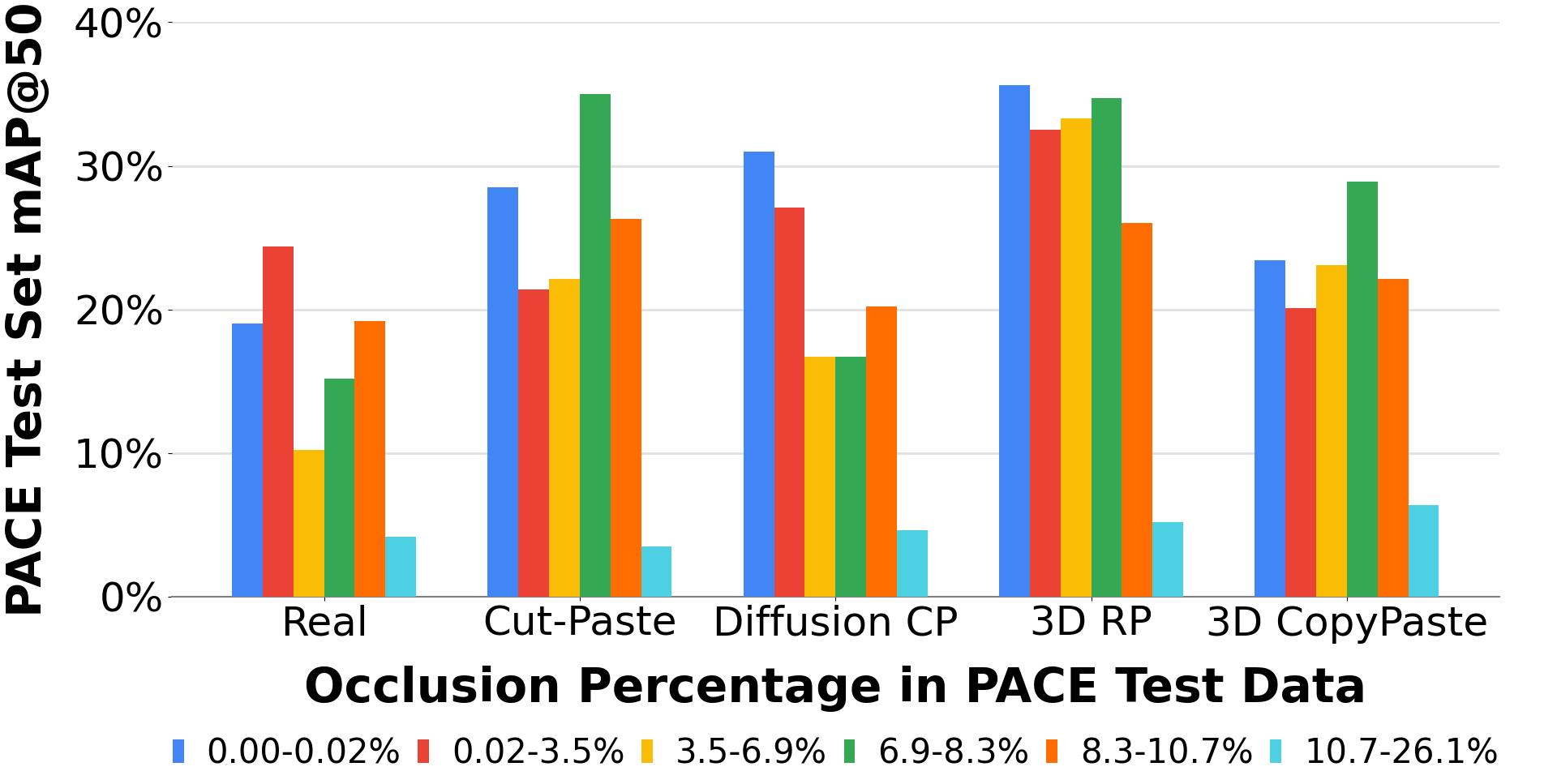}
    \caption{PACE test set performance stratified by occlusion percentage (i.e., the total overlap area divided by the total area of all foreground objects).}
    \label{fig:occlusion}
\end{figure}

We investigate the ability of our data synthesis methods to aid in detecting occluded objects in \cref{fig:occlusion}. The occlusion percentage, calculated as the total overlap area divided by the total area of all foreground objects, is binned into six categories, ranging from 0\% to roughly 25\% occlusion. We find that our 3D Random Placement method achieves the best results on average across all bins by this metric, followed by the strong performance from the Cut-Paste method. \Cref{fig:occlusion} summarizes our results.

\subsection{Data Mixing}\label{subsec:datamixing}
We also examined the effect of different data mixing schemes, including combining multiple types of synthetic data to increase dataset diversity. As expected, the runs mixed with the real data performed the best, with a dominant performance from the Diffusion Copy-Paste method (see \Cref{tab:map_matrix}). Aside from these runs, however, we observed that mixing a synthetic data type with another synthetic method improved performance by an average of 3.2\% mAP for Cut-Paste, 0.36\% mAP for 3D Random Placement, 3.57\% mAP for Diffusion Copy-Paste, and 12.1\% mAP for 3D Copy-Paste. This suggests that combining multiple data augmentation methods can yield substantial performance boosts and compensate for deficiencies from any particular method.
\begin{table}[ht]
\centering
\begin{tabular}{l|ccccc}
\toprule
\textbf{Mix with} & \textbf{Real} & \textbf{C-P} & \textbf{3DRP} & \textbf{DiffCP} & \textbf{3DCP} \\
\midrule
\textbf{Real}         & 37.4\% & 34.9\% & 42.2\% & 50.4\% & 44.5\% \\
\textbf{C-P}        & -- & 34.9\% & 41.9\% & 36.9\% & 35.5\% \\
\textbf{3DRP}         & -- & -- & 43.6\% & 45.7\% & 44.3\% \\
\textbf{DiffCP}       & -- & -- & -- & 39.5\% & 46.6\% \\
\textbf{3DCP}         & -- & -- & -- & -- & 30.0\% \\
\bottomrule
\end{tabular}
\caption{mAP results for models trained on individual datasets (diagonal) and pairwise combinations (off-diagonal). Here C-P stands for Cut-Paste, 3DRP for 3D Random Placement, DiffCP for Diffusion Copy-Paste, and 3DCP for 3D Copy-Paste.}\label{tab:map_matrix}
\end{table}

\section{Discussion}\label{sec:conclusion}

In this paper, we discussed different methods of generating synthetic data for improving object detection in data-limited settings. We compared methods using both 2D and 3D object information, as well as methods with random versus contextually guided object placement. Our specific experimental setup involving data scarcity and occlusion-modeling, both motivated by constraints arising from a practical application of object detection, lead us to apply and develop two data synthesis methods from the literature (Cut-Paste and 3D Copy-Paste) and two of our own (Diffusion Copy-Paste and 3D Random Placement) tailored to these challenges. Our results demonstrate the considerable power of synthetic data augmentation in this data-limited regime, and particularly showed that our developed methods achieve strong performance boosts over widely-used and relevant data synthesis methods in the literature, such as Cut-Paste. 

Comparison across four methods reveals mixed results for when modeling visual context and realism can lead to improved performance. 3D Random Placement and Diffusion Copy-Paste produce more realistic datasets than Cut-Paste by incorporating geometric, lighting, \& perspective information, and contextual background information respectively, and both outperform the Cut-Paste baseline. 3D Copy-Paste combines aspects of both but fails to surpass Cut-Paste. We note that adding certain \textit{realistic} constraints (e.g. placing objects on surfaces) can unintentionally \textit{reduce diversity} along other aspects, such as viewpoint coverage, which may counteract the advantages of modeling the real world more faithfully.  We hope our work invites further empirical study of the under-addressed yet practical object-centric setting.
\section*{Acknowledgements}
This work was supported by the Institute for Pure and Applied Mathematics (IPAM) at UCLA, in conjunction with the industry sponsor Analog Devices, Inc. (ADI), through the 2025 Research in Industrial Projects for Students (RIPS) program. We would like to thank Dr. Susana Serna, the RIPS Program Director, for her support and feedback throughout the program. 
{
    \small
    \bibliographystyle{ieeenat_fullname}
    \bibliography{main}
}

\end{document}